\documentclass[10pt,twocolumn,letterpaper]{article}

\usepackage{cvpr}              

\usepackage{graphicx}
\usepackage{amsmath}
\usepackage{amssymb}
\usepackage{booktabs}
\usepackage{multirow}
\usepackage{float}
\usepackage{caption}

\usepackage[pagebackref,breaklinks,colorlinks]{hyperref}

\usepackage[capitalize]{cleveref}
\crefname{section}{Sec.}{Secs.}
\Crefname{section}{Section}{Sections}
\Crefname{table}{Table}{Tables}
\crefname{table}{Tab.}{Tabs.}

\def\cvprPaperID{121} 
\def\confName{CVPR}
\def\confYear{2023}

\begin{document}

\title{Face Animation with an Attribute-Guided Diffusion Model}

\author{
Bohan Zeng{$^{1}$}\thanks{These authors contributed equally.}, Xuhui Liu{$^1$}\footnotemark[1], Sicheng Gao{$^{1}$}\footnotemark[1], Boyu Liu{$^{1}$}, Hong Li{$^{1}$}\\
Jianzhuang Liu{$^2$}, Baochang Zhang{$^{1,3}$}\thanks{Corresponding Author: bczhang@buaa.edu.cn.} \\ 
{$^1$}Beihang University \\ {$^2$}Shenzhen Institutes of Advanced Technology, University of Chinese Academy of Sciences \\ {$^3$}Zhongguancun Laboratory, Beijing, China \\
}
\maketitle


\begin{abstract}
Face animation has achieved much progress in computer vision. However, prevailing GAN-based methods suffer from unnatural distortions and artifacts due to sophisticated motion deformation. In this paper, we propose a Face Animation framework with an attribute-guided Diffusion Model (FADM), which is the first work to exploit the superior modeling capacity of diffusion models for photo-realistic talking-head generation. To mitigate the uncontrollable synthesis effect of the diffusion model, we design an Attribute-Guided Conditioning Network (AGCN) to adaptively combine the coarse animation features and 3D face reconstruction results, which can incorporate appearance and motion conditions into the diffusion process. These specific designs help FADM rectify unnatural artifacts and distortions, and also enrich high-fidelity facial details through iterative diffusion refinements with accurate animation attributes. FADM can flexibly and effectively improve existing animation videos. Extensive experiments on widely used talking-head benchmarks validate the effectiveness of FADM over prior arts. 
The source code is available in \url{https://github.com/zengbohan0217/FADM}.

\end{abstract}

\vspace{-5mm}
\section{Introduction}
\label{introduction}

Face animation, referring to the task of animating a still face with poses and expressions provided by a driving video, has drawn increasing attention due to its wide application scenarios, such as photography, online conferencing, social media, and video production. With the progress of generative models such as Generative Adversarial Networks (GANs), recent face animation methods have achieved impressive performance in synthesizing high-fidelity talking faces. However, they still suffer from undesirable artifacts and distortions in generated results.

Existing face animation methods are mostly based on GAN models, which mainly divide the generation process into warping and rendering. They utilize the difference of expressions and poses between the source image and the driving video to calculate the motion flow, which can guide the further warping process of the encoded source features. After that,  the warped features are fed into a decoding module for rendering and synthesizing the final results.
These methods can be roughly classified into three categories: model-free \cite{siarohin2019animating, siarohin2019first, wang2022latent, burkov2020neural}, landmark-based \cite{zakharov2020fast,  wang2019few,  zakharov2019few} and 3D structure-based \cite{koujan2020head2head, doukas2021headgan, wang2021one}. They obtain promising performances in preserving the identity and appearance of the source and generate relatively accurate motion from the driving video. 
However, due to the restricted ability of adversarial learning on high-fidelity appearance reconstruction, these GAN-based methods focus more on face distributions but not much on facial details, and thus they might generate unnatural artifacts and distortions (see Fig. \ref{intro_pic}).

Recently, the great success of denoising diffusion probabilistic models (DMs) in computer vision, such as in-painting \cite{li2022srdiff,  saharia2022image, rombach2022high, lugmayr2022repaint, saharia2022palette, ho2022cascaded, chung2022score,  song2022MedicalImaging}, video synthesis \cite{harvey2022flexible, ho2022video, yang2022diffusion, zhang2022motiondiffuse}, and 3D points cloud modeling \cite{Zhou2021PVD, luo2021diffusion,  lyu2022DiffusionRefinement}, indicates their superior capacity in generative tasks. They can model highly complex data distributions through a sequence of diffusion refinement steps. Based on optimizing a variant of the variational lower bound, diffusion models can effectively avoid the distortion problem encountered by GANs and generate high-fidelity facial details. However, existing DMs tend to encode images into arbitrarily high-variance latent spaces without specific attribute restrictions, which is unqualified for face animation that has explicit requirements on the facial appearance, pose, and expression.

In this paper, we enable face animation with an iterative denoising diffusion process to rectify the distortions and unnatural artifacts, while ensuring accurate animation attributes. We propose a Face Animation framework with an attribute-guided Diffusion Model (FADM) for photo-realistic talking-head generation. 
Specifically, we introduce a Coarse Generative Module (CGM) to obtain the preliminary animation results, which provide low-resolution features for the diffusion process. To mitigate the high variability of DMs, we design an Attribute-Guided Conditioning Network (AGCN) to incorporate appearance and motion conditions into the iterative refinement process. On the one hand, we utilize an encoder network to extract the appearance code from the driving frames and coarse features and introduce an MSE loss to align them. On the other hand, we leverage a 3D reconstruction module to predict the poses and expressions of the source and driving frames. Based on these, AGCN uses a Multi-Layer Perceptron (MLP) to assign different confidence values for the multi-resolution features, to adaptively adjust the expressed ratio of the coarse features and fuse them effectively as motion condition. Therefore, the diffusion refined process is well guided to synthesize accurate and fine-grained talking-head videos, as shown in Fig. \ref{intro_pic}.
Moreover, it is worth noting that FADM can also be directly applied to improving the quality of existing animation videos as a flexible talking-head rectification tool.
The contributions of this paper are summarized as follows:

\begin{itemize}
    \item We propose a Face Animation framework with an attribute-guided Diffusion Model (FADM) to rectify the distortions and unnatural artifacts, which can also enrich the facial details through an iterative diffusion refinement process. 

    \item We design an Attribute-Guided Conditioning Network (AGCN) to adaptively extract appearance and motion conditions for the diffusion process and ensure the validity of generated results. Moreover, FADM can flexibly and effectively improve the quality of available animation videos.

    \item Extensive experiments are conducted to compare FADM with state-of-the-art methods. The results show that FADM generates overall best qualitative and quantitative results on widely used talking-head benchmarks, and genuinely achieves photo-realistic face animation.
    
\end{itemize}

\section{Related Work}
\label{related work}

\subsection{GAN-based Face Animation Models}
Model-free methods \cite{wiles2018x2face, siarohin2019animating, siarohin2019first, wang2022latent, burkov2020neural, wang2021one} learns the motion field for the deformation of face images in a self-supervised manner without additional facial priors.  MonkeyNet \cite{siarohin2019animating} predicts sparse key-points to complete motion transfer.
In particular, the First Order Motion Model (FOMM) \cite{siarohin2019first} significantly improves the performance of face animation with a rigorous first-order mathematical model. 
Face vid2vid \cite{wang2021one} extends FOMM by introducing 3D representations and achieves realistic face animation. Nevertheless, it has a considerable computational cost and performs poorly in expression transformation.
Landmark-based methods \cite{averbuch2017bringing, geng2018warp, zakharov2020fast, ha2020marionette, wang2019few, zakharov2019few} utilize 2D facial landmarks as conditions for reenactment. However, these methods often cannot handle identity preservation well during the generation process.
More recently, many 3D structure-based works \cite{kim2018deep,   kim2019neural, koujan2020head2head, doukas2021headgan, ren2021pirenderer, wang2021safa, zeng2022fnevr}  resort to the geometric prior of 3D faces and achieve impressive results on realistic talk-head synthesis.
HeadGAN \cite{doukas2021headgan} takes the rendered 3D mesh as input and predicts the depth to deform the face, but it fails in expression transferring. Conditioned on the parameters of the 3D Morphable Model (3DMM) \cite{blanz1999morphable}, StyleRig \cite{tewari2020stylerig} and GIF \cite{ghosh2020gif} respectively employ pre-trained StyleGAN \cite{karras2019style} and StyleGAN2 \cite{karras2020analyzing} to warp face images. PIRenderer \cite{ren2021pirenderer} controls the face motions and predicts a flow field for deformation. And FNeVR \cite{zeng2022fnevr}, AD-NeRF \cite{guo2021ad} adopt the NeRF \cite{mildenhall2020nerf} to generate high-quality face animation.

Unfortunately, all of these methods rely on the GAN models to generate animated faces, which also brings unnatural distortions and artifacts. This paper enables current face animation methods with diffusion refinements to achieve high-fidelity face animation. 

\begin{figure*}[t]
\centering
\includegraphics[width=0.78\textwidth]{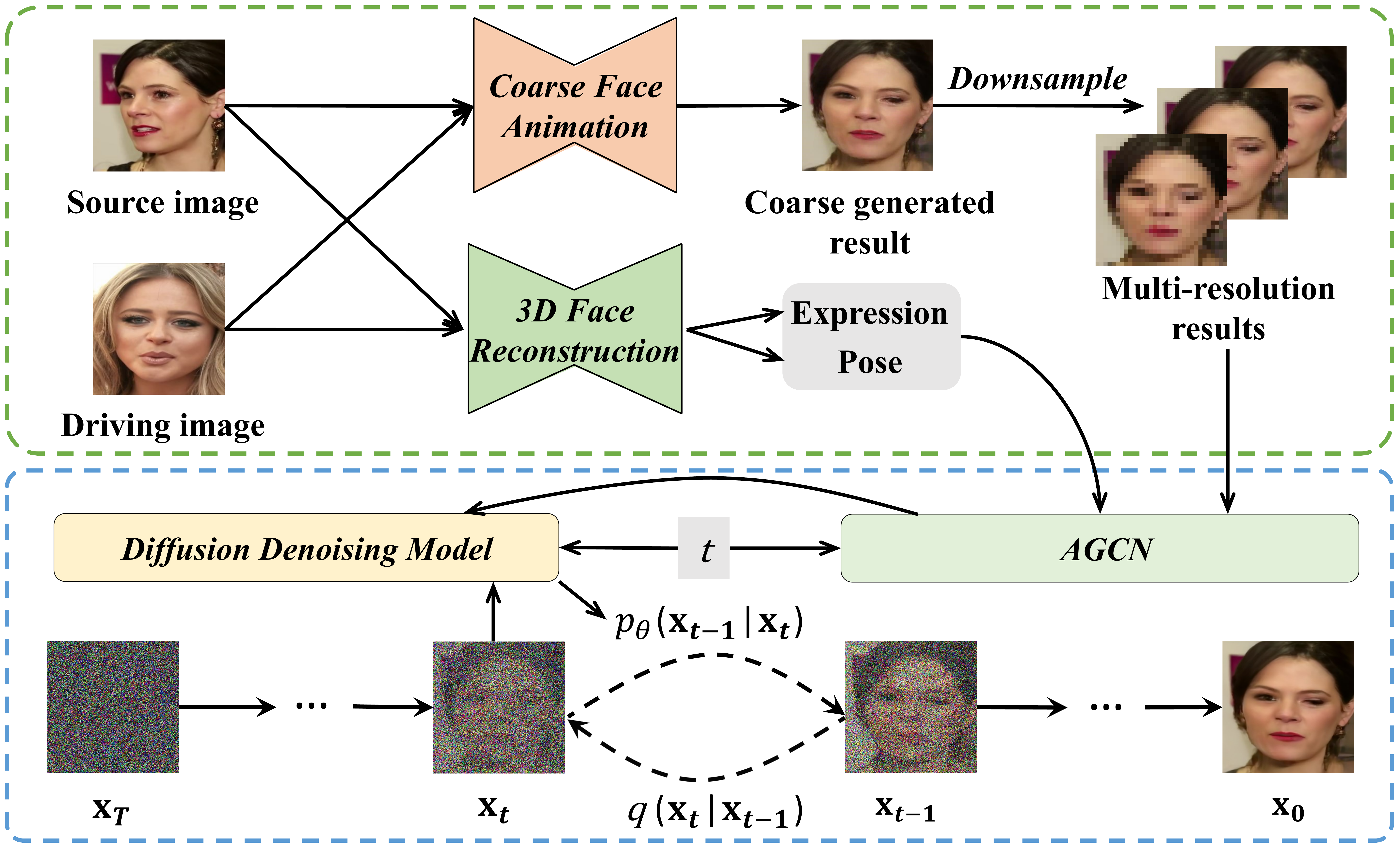} 
\caption{FADM framework. It consists of the coarse face animation generator, the encoder of the 3D face reconstruction model DECA \cite{feng2021learning}, and the core diffusion rendering module (blue box). We first obtain the coarsely generated result and the facial expression and pose information, on which AGCN is then performed to estimate the appearance and motion conditions for further rendering by the diffusion model.}
\label{framework overview}
\end{figure*}

\subsection{Diffusion Probabilistic Models}

Diffusion Probabilistic Models (DMs) are first presented in \cite{Sohl-DicksteinW2015Deep} as a kind of generative models, which use a Markov chain to gradually add noise to obtain a latent variable, and then gradually transform the latent variable to obtain the generated results through a learned iterative denoising process. 
Recently, DMs have achieved state-of-the-art results in various synthesis tasks, including image synthesis \cite{ho2020denoising, saharia2022palette, ho2022cascaded}, speech synthesis \cite{kong2020diffwave, liu2021diffsinger},  and 3D cloud related tasks \cite{Zhou2021PVD, luo2021diffusion, lyu2022DiffusionRefinement}. In image synthesis,
\cite{ho2020denoising} and \cite{song2020denoising} show the superior capability of diffusion models to generate high-quality images in many computer vision tasks.
For example, \cite{li2022srdiff, saharia2022image, rombach2022high, lugmayr2022repaint, saharia2022palette, ho2022cascaded, chung2022score, song2022MedicalImaging, gao2023implicit} exhibit impressive performance on image super-resolution and in-painting. Particularly, the stable diffusion model \cite{rombach2022high} has been applied to various practical application scenarios, such as text-image generation, and has given rise to a wave of DMs. Likewise, the amazing performance of DMs in semantic segmentation \cite{baranchuk2022LabelEfficient}, point cloud completion and generation \cite{Zhou2021PVD, luo2021diffusion, lyu2022DiffusionRefinement}, and video generation \cite{harvey2022flexible, ho2022video,  yang2022diffusion,  zhang2022motiondiffuse} again demonstrates their excellent capabilities.

However, given that current DMs mostly have no strict requirements on the attributes of the generated results, the results often lie in an arbitrarily high-variance space. In contrast, face animation strictly demands animating the source with explicit poses and expressions provided by the driving video, while preserving the appearance of the source. 



\section{Method}
Since GANs have limited capability to model complex facial structures and motion for our task, existing methods often suffer from essential distortions and unnatural artifacts. To address this problem, we propose the Face Animation framework with a Diffusion Model (FADM), which is comprised of: (1) a coarse generative module, (2) a 3D face reconstruction model, (3) an attribute-guided conditioning network, and (4) a diffusion rendering module. The overview of FADM is shown in Fig. \ref{framework overview}. In this section, we describe the details of FADM and elaborate on how it rectifies current face animations with explicit and accurate attributes through a sequence of diffusion refinement steps. 


\subsection{Coarse Generative Module}
In FADM, we first use a Coarse Generative Module (CGM) such and FOMM \cite{siarohin2019first} or Face vid2vid \cite{wang2021one} to generate coarse animation images. 
Given a source image $s$ and the driving frames $d$, the objective of CGM is to deform $s$ with the expression and pose information derived from $d$, while keeping the appearance information of $s$. 
It includes two steps: warping the source features first according to the expressions and poses of the driving frames, and then rendering warped features to obtain final animation images. Intuitively, the general process of generating the coarse animation results $g$ can be conducted as:
\begin{equation}
\small
    \begin{aligned}
        g = G({\rm Warp}(s, exp_d, pose_d)), \\
    \end{aligned}
    \label{coarse generation}
\end{equation}
where $G$ represents the generative model, and $exp_d$ and $pose_d$ denote the expressions and poses of the driving frames, respectively. Although the coarse results have a promising performance in preserving the appearance of $s$ and transferring motion from $d$, there often exist undesirable distortions in the facial details and the background area beyond the face. To alleviate this problem, we leverage a diffusion process with explicit conditions to renovate the coarse results.

\begin{figure}[t]
\centering
\includegraphics[width=0.5\textwidth]{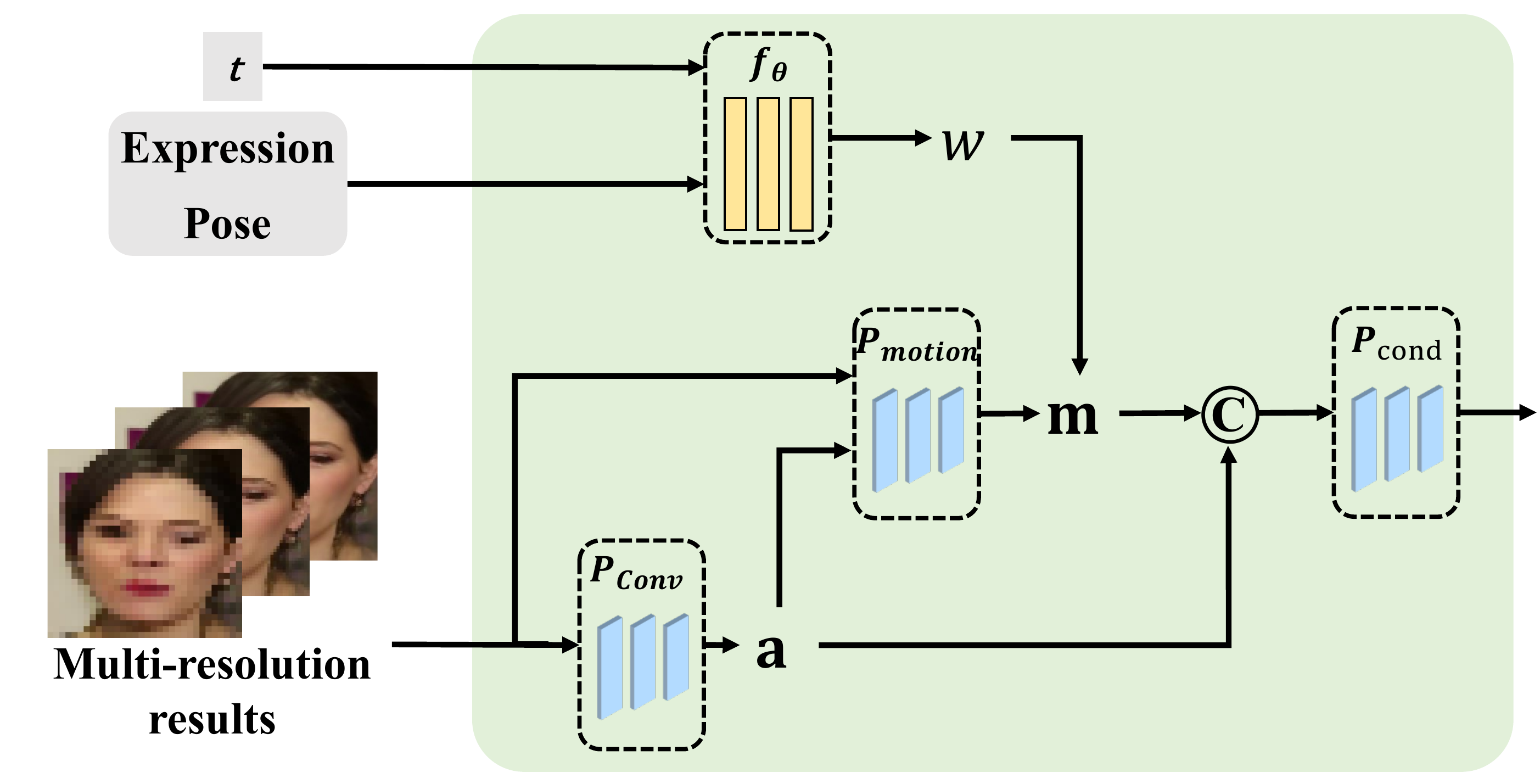} 
\caption{Architecture of AGCN during inference, where $\copyright$ indicates the channel-wise concatenation. }
\label{AGCN fig}
\end{figure}
\vspace{-1mm}

\subsection{Diffusion Rendering Module}
To handle the distortion problem caused by CGM, a diffusion rendering module (the blue box in Fig. \ref{framework overview}) is designed in FADM to synthesize photo-realistic images through an iterative diffusion process from coarse to fine. Here we first give the preliminaries of DMs, and then describe this module for face animation.
\vspace{-3mm}
\subsubsection{Preliminaries of Diffusion Models}

Following \cite{ho2020denoising}, we define the inference process $p_{\theta}$ of DMs, which denoises a normally distributed variable $x_{T}$ to a target image $x_{0}$ as:
\vspace{-1mm}
\begin{equation}
\small
    \begin{aligned}
        p_{\theta}(\mathbf{x}_{0:T}) = & p(\mathbf{x}_{T}) \prod^{T}_{t=1} p_{\theta}(\mathbf{x}_{t-1} | \mathbf{x}_t),  \\
        p(\mathbf{x}_{T}) = & \mathcal{N}\left(\mathbf{x}_T \mid {0}, {I}\right), \\
        p_{\theta}(\mathbf{x}_{t-1} | x_t) = & \mathcal{N}(\mathbf{x}_{t-1};\mu_{\theta}(\mathbf{x}_t, t), \Sigma_{\theta}(\mathbf{x}_t, t)),
    \end{aligned}
\end{equation}
where $\mathbf{x}_1, ..., \mathbf{x}_T$ are latent features with added noise, $p_{\theta}(\mathbf{x}_{0:T})$  represents the joint distribution which performs the image generation process and is defined as a Markov chain with learnable Gaussian transitions $p_{\theta}(\mathbf{x}_{t-1}|\mathbf{x}_t)$. Inversed to the inference process, the forward process gradually adds Gaussian noise to $\mathbf{x}_{0}$ over $T$ iterations, which can be expressed as:
\vspace{-1mm}
\begin{equation}
\small
    \begin{aligned}
        q(\mathbf{x}_{1:T} | \mathbf{x}_0) = & \prod^{T}_{t=1} q(\mathbf{x}_t | \mathbf{x}_{t-1}),  \\
        q(\mathbf{x}_t | \mathbf{x}_{t-1}) = & \mathcal{N}(\mathbf{x}_{t}; \sqrt{1-\beta_t} \mathbf{x}_{t-1}, \beta_t \bf{I}),
    \end{aligned}
\end{equation}
where $\beta_1, ..., \beta_T$ are the variance schedule. By optimizing the negative log-likelihood of the data distribution through the variational lower bound, the optimization objective can be interpreted as learning an equally weighted sequence of a denoising model $z_{\theta}(\mathbf{x}_t, t), t \in \{1, 2, ..., T\}$:
\begin{equation}
\small
    \begin{aligned}
        \mathcal{L}_{\theta} =& E_{\mathbf{x}_0, z \sim \mathcal{N}(0,1), t}[\parallel  z - z_{\theta}(\mathbf{x}_t, t) \parallel^2_2].
    \end{aligned}
\end{equation}


\subsubsection{Diffusion Rendering for Face Animation}

In order to avoid the mismatch between the DM's high-variance encoding and the explicit attribute requirements of face animation, the DM needs to generate the rendering face frames in strict conformity with the appearance of the source, having the poses and expressions of the driving frames.
Accordingly, we formulate the inference process of our diffusion rendering model as:
\begin{equation}
\small
    \begin{aligned}
        p_{d}(\mathbf{x}_{0:T}) = & p(\mathbf{x}_{T}) \prod^{T}_{t=1} p_{d}(\mathbf{x}_{t-1} | \mathbf{x}_t, \mathbf{a}, \mathbf{m}),  \\
        p_{d}(\mathbf{x}_{t-1} | \mathbf{x}_t, \mathbf{a}, \mathbf{m}) = & \mathcal{N}(\mathbf{x}_{t-1};\mu_{d}(\mathbf{x}_t, t, \mathbf{a}, \mathbf{m}), \Sigma_{d}(\mathbf{x}_t, t, \mathbf{a}, \mathbf{m})),
    \end{aligned}
\end{equation}
where $\mathbf{a}$ donates the appearance code of the source image, and $\mathbf{m}$ donates the motion condition derived from the variation between the source image and driving frames' poses and expressions.
Consequently, the optimization objective of our diffusion rendering model is defined as a conditional diffusion loss $\mathcal{L}_{d}$:
\begin{equation}
\small
  \begin{aligned}
  \mathcal{L}_{d} =& E_{\mathbf{x}_0, (\mathbf{a}, \mathbf{m}), z \sim \mathcal{N}(0,1), t}[\parallel  z - z_{d}(\mathbf{x}_t, t, (\mathbf{a}, \mathbf{m})) \parallel^2_2]. \\
  \end{aligned}
  \label{base diffusion loss}
\end{equation}

To minimize $\mathcal{L}_{d}$ with $(\mathbf{a}, \mathbf{m})$, we design an Attribute-Guided Conditioning Network (AGCN) to extract appropriate appearance and motion conditions, and fuse them adaptively for navigating the diffusion process, as shown in Fig. \ref{AGCN fig}. 
The appearance condition is used to provide the diffusion process with faithful characteristics of the source image, while the motion condition can impose restriction on the generated poses and expressions and dynamically modify the facial details. 


\paragraph{Appearance Condition.}
Considering the current training style of using the same identity of the source and the driving frames, we note that the driving frames are the most appropriate conditions to provide faithful appearance information for the subsequent diffusion process, while only the coarse animation results are available during the inference process, which are the sub-optimal choice. Formally, we design a CNN encoder $P_{\operatorname{Conv}}$  to extract the appearance code $\mathbf{a}$:
\begin{equation}
\small
  \mathbf{a}=\left\{
  \begin{aligned}
  & P_{\operatorname{Conv}}(\downarrow_*(d)), \quad {\rm in \ training} \\
  & P_{\operatorname{Conv}}(\downarrow_*(g)), \quad {\rm in \ inference}, \\
  \end{aligned}
  \right.
  \label{color condition}
\end{equation}
where $\downarrow_*$ denotes the downsampling operation. As stated above, since the coarse animation results might involve unexpected interference with the appearance, we aim to alleviate the interference and guarantee the creditability of $\mathbf{a}$ provided during inference. Specifically, we respectively input the driving frames and coarse animation results into $P_{\operatorname{Conv}}$, and then align the appearance conditions predicted from $d$ and $g$ through an MSE loss $\mathcal{L}_{color}$ as:
\begin{equation}
\small
    \begin{aligned}
        \mathcal{L}_{color} = & {\rm MSE} (P_{\operatorname{Conv}}(\downarrow_*(d)), P_{\operatorname{Conv}}(\downarrow_*(g))). \\
    \end{aligned}
\end{equation}
Here $d$ and $g$ are both fixed features irrelevant to $t$. $\mathcal{L}_{color}$ works by facilitating $P_{\operatorname{Conv}}$ to extract the most valuable appearance information from $g$. Consequently, $P_{\operatorname{Conv}}$ is capable of providing faithful appearance conditions during the inference, which are comparable with those provided by $d$ in the training process.
\vspace{-1mm}
\paragraph{Motion Condition.}
Considering FADM aims to improve the quality of the coarse animation results, taking them as the motion condition seems to be an intuitively effective choice for the diffusion process. Nonetheless, they may also bring distortions in the diffusion process. Empirically, the coarse results suffer from extreme distortions when the motion changes dramatically between the source and the driving frames. In this case, features with a higher resolution tend to contain more distortions, while features with a lower resolution could weaken them. In other words, compared with high-resolution features, low-resolution features allow the diffusion process to synthesize richer facial details to compensate for the distortions. Based on this observation, we handle this problem in an adaptive way, seeking a balance in fusing multi-resolution coarse animation results as the motion condition, to alleviate the distortions and enrich the facial details on the basis of ensuring accurate animation attributes.

Specifically, we first utilize the downsampling operation to process the coarse animation result into three coarse animation features with different resolutions. 
Meanwhile, we exploit the advanced 3D face reconstruction model DECA \cite{feng2021learning}  to extract the facial poses $pose$ and expressions $exp$ from the source and the driving frames, and concatenate them as the motion state. Then, an MLP $f_{\theta}$ is introduced as a motion measuring function to model the changing amplitude of the motion between the source image and the driving frames, so as to obtain the motion weight $w$. The process (Fig. \ref{AGCN fig}) is represented as :
\begin{equation}
\small
  \begin{aligned}
  w =& f_{\theta}({\rm Concat}(exp_s, pose_s) - {\rm Concat}(exp_d, pose_d), t), \\
  \end{aligned}
  \label{motion weight}
\end{equation}
where $t$ is an arbitrary timestep of the diffusion process. 


With the initial weight, we assign a larger value to the features with a lower resolution when the motion changes drastically. If the motion does not change much, the features with a higher resolution should be allocated greater weights for guaranteeing high-fidelity generation. Formally, we calculate the motion condition $\mathbf{m}$ by:
\begin{equation}
\small
    \begin{aligned}
        w_i = & \frac{K-i}{K} \cdot \exp(w - \alpha) + \frac{i}{K} \cdot \exp(-w + \alpha),  \\
        \mathbf{m} = & \sum^{K}_{i=1} w_i \cdot P_{motion}(g_i, \mathbf{a}),
    \end{aligned}
    \label{motion condition}
\end{equation}
where $\alpha$ is a hyper-parameter,  $g_i$ and $w_i$ denote the coarse generated images and the motion weights for refined images with different resolutions, respectively, $K$ is the number of resolutions, and $P_{motion}$ is a CNN to generate the motion conditions in different resolutions. Note that we set $\alpha=0.3$ in this paper.

Lastly, we introduce a CNN $P_{cond}$ to fuse the appearance condition $\mathbf{a}$ and the motion condition $\mathbf{m}$ together.
In general, the objective of our diffusion model is rewritten as:
\begin{equation}
\small
  \begin{aligned}
  \mathcal{L}_{d} =& E_{\mathbf{x}_0, (\mathbf{a}, \mathbf{m}), z \sim \mathcal{N}(0,1), t}[\parallel  z - z_{d}(\mathbf{x}_t, t, P_{cond}(\mathbf{a}, \mathbf{m}, t)) \parallel^2_2], \\
  \end{aligned}
  \label{final loss}
\end{equation}
where $z_d$ denotes the diffusion denoising model. Following \cite{ song2020denoising}, we employ an U-net architecture as the denoising model which is optimized to iteratively remove the noise, and synthesize high-fidelity target faces in 100 timesteps.

It is worth noting that our FADM can be directly used to improve the visual quality of existing animated videos by setting the first frame as the source image, bringing great convenience in practice. Related analysis and visual results are provided in the supplementary materials.


\begin{table*}[t]
  \caption{Quantitative comparison of same-identity reconstruction on VoxCeleb \cite{nagrani2017voxceleb}.}
  \begin{center}
  \setlength{\tabcolsep}{2mm}{
  \label{reconstruction}
  \centering
  \renewcommand{\arraystretch}{1.2}

  \begin{tabular}{lcccccc}
    \hline
    Method      & $\mathcal{L}_1$ $\downarrow$   & LPIPS $\downarrow$   & PSNR $\uparrow$  & SSIM $\uparrow$  & AKD $\downarrow$  & AED $\downarrow$   \\
    \hline
    Bilayer \cite{zakharov2020fast}   & 0.1753   & 0.5733  & 12.802   & 0.3201   & 13.83  & 0.0564     \\
    PIRender \cite{ren2021pirenderer} & 0.0574   & 0.2225  & 21.154   & 0.6564   & 2.249  & 0.0321     \\
    FOMM \cite{siarohin2019first}     & 0.0451   & 0.1479  & 23.422   & 0.7521   & 1.456  & 0.0247    \\
    Face vid2vid \cite{wang2021one}   & 0.0456   & 0.1395  & 23.279   & 0.7487   & 1.615  & 0.0258     \\
    DaGAN \cite{hong2022depth}        & 0.0468   & 0.1465  & 23.449   & 0.7564   & 1.546  & 0.0257    \\
    \hline
    FADM           & \textbf{0.0402}   & \textbf{0.1379}  & \textbf{24.434}   & \textbf{0.7841}   & \textbf{1.392}  & \textbf{0.0241}     \\
    \hline
  \end{tabular}
  }
  \end{center}
\end{table*}

\begin{figure*}[t]
\centering
\includegraphics[width=0.85\textwidth]{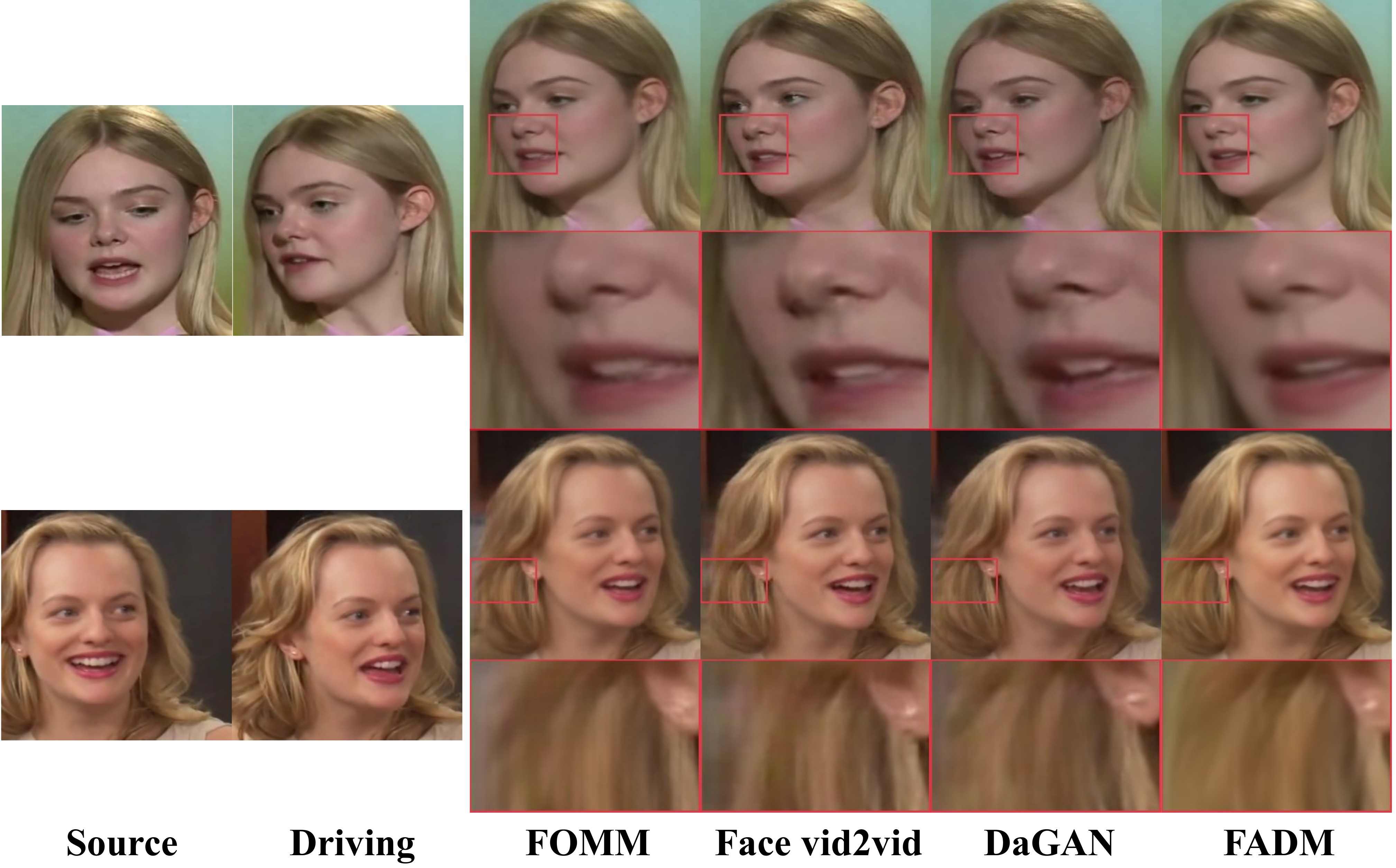} 
\caption{Qualitative comparison with SOTA methods on the reconstruction task. Evidently, our FADM can produce more fine-grained details, and is effective to rectify the unnatural parts. }
\label{reconstruction fig}
\end{figure*}

\section{Experiments}

\subsection{Implementation Details}

\paragraph{Datasets.}
We evaluate the performance of FADM on three datasets: VoxCeleb \cite{nagrani2017voxceleb}, VoxCeleb2 \cite{chung2018voxceleb2}, and CelebA \cite{liu2018large}. VoxCeleb contains about 100,000 videos covering 1,251 different speakers. VoxCeleb2 has more than 1M videos of different celebrities. CelebA consists of 200,000 images of 10,000 different persons with different genders and multi-age groups. Note that we use the images from CelebA as the source images to evaluate the performance on the reenactment task.
Following FOMM \cite{siarohin2019first}, we preprocess the data by cropping faces from the videos and resizing them to 256×256.
\vspace{-1mm}
\paragraph{Training Details.}
We use the pretrained FOMM \cite{siarohin2019first} or Face vid2vid \cite{wang2021one} to generate the coarse face animation results and then train FADM for about 100 epochs with the images from the videos repeating 75 times per epoch. We adopt the Adam \cite{kingma2014adam} optimizer with learning rate $\eta = 2 \times 10^{-4}$, $\gamma_1 = 0.5$ and $\gamma_2 = 0.9$. Furthermore, we use four 24GB NVIDIA 3090 GPUs for training.
\vspace{-1mm}
\paragraph{Evalution Metrics.}
The evaluation metrics include: (1) $\mathcal{L}_1$, PSNR, and SSIM \cite{ZhouWang2004ImageQA} 
(2) LPIPS \cite{zhang2018perceptual} and FID \cite{MartinHeusel2017GANsTB} 
(3) Average Keypoint Distance (AKD) and Average Euclidean Distance (AED) as set in \cite{siarohin2019first}; 
(4) identity preservation cosine similarity (CSIM) \cite{richardson2021encoding, huang2020curricularface} calculated by CircularFace \cite{huang2020curricularface} 

\begin{figure*}[t]
\centering
\includegraphics[width=0.85\textwidth]{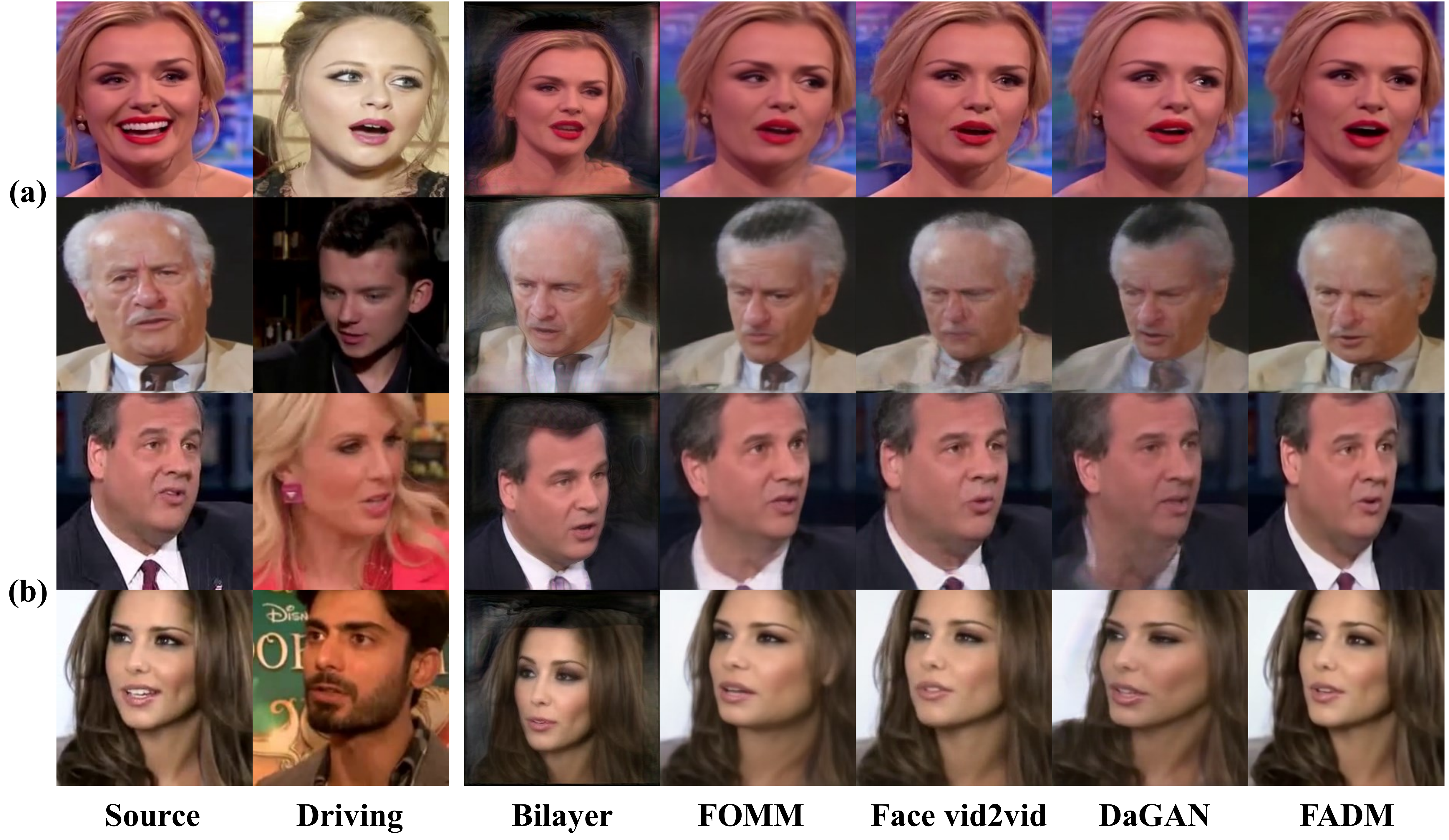} 
\caption{Visual comparison with SOTA methods on the reenactment task. \textbf{(a)} Results on VoxCeleb \cite{nagrani2017voxceleb}. \textbf{(b)} Results on VoxCeleb2 \cite{chung2018voxceleb2}. On both testing datasets, our FADM can produce more identity-preserving and photo-realistic results compared with other methods. Particularly, compared with the most outstanding existing method Face vid2vid, FADM can effectively alleviate the unnatural artifacts on the generated face images.}
\label{reenacment fig}
\end{figure*}

\begin{table*}[t]
    \caption{Quantitative comparison for cross-identity reenactment on the testing datasets of VoxCeleb \cite{nagrani2017voxceleb}, VoxCeleb2 \cite{chung2018voxceleb2}, and CelebA \cite{liu2018large}.}
    \centering
    \setlength{\tabcolsep}{2mm}{
    \label{reenactment}
    \renewcommand{\arraystretch}{1.2}
    \begin{tabular}{l|cc|cc|cc}
        \hline
        \multirow{2}{*}{Method}&    \multicolumn{2}{|c}{VoxCeleb}&    \multicolumn{2}{|c}{VoxCeleb2}&   \multicolumn{2}{|c}{CelebA}\\\cline{2-7} 
                      & FID$\downarrow$          & CSIM $\uparrow$      & FID$\downarrow$          & CSIM $\uparrow$     & FID$\downarrow$          & CSIM $\uparrow$\\
        \hline
        FOMM              & 106.9   & 0.5491   & \textbf{138.1}   & 0.5228   & 96.29   & 0.5410     \\ 
        Face vid2vid    & \textbf{106.6}   & 0.6447   & 148.6   & 0.6290   & 93.44   & 0.6218     \\
        DaGAN             & 110.3   & 0.5305   & 139.6   & 0.4932   & 96.47   & 0.4983     \\
        \hline   
        FADM              & \textbf{106.6}   & \textbf{0.6598}   & 151.7   & \textbf{0.6320}   & \textbf{86.55}   & \textbf{0.6366}    \\ 
        \hline
      \end{tabular}}
\end{table*}

\subsection{Comparison with State-of-the-Art Methods}
\vspace{-1mm}
\paragraph{Methods.}
We compare our FADM with five state-of-the-art methods: FOMM \cite{siarohin2019first}, Face vid2vid \cite{wang2021one}, Bilayer \cite{zakharov2020fast}, DaGAN \cite{hong2022depth}, and PIRenderer \cite{ren2021pirenderer}. For Bilayer, FOMM, DaGAN, and PIRenderer, we use their official pre-trained models for evaluation, while for Face vid2vid, we adopt a widely recognized unofficial model due to the absence of the official code. All of these models are pre-trained on VoxCeleb.
\vspace{-2mm}

\paragraph{Same-Identity Reconstruction.}

We conduct quantitative and qualitative comparisons of the same-identity reconstruction task on the VoxCeleb dataset, where FOMM \cite{siarohin2019first} is used to generate the coarse face animation results.
In order to accelerate the inference, we randomly select several short-time videos from the testing dataset of VoxCeleb for quantitative comparison, and the selected video list is available in the supplementary materials. 
As shown in Table \ref{reconstruction}, compared to other SOTA methods, it is evident that FADM achieves the best performance in all metrics, especially the reconstruction faithfulness $\mathcal{L}_1$ and the visual quality LPIPS, demonstrating the superiority of our FADM in generating high-fidelity face animation.
Moreover, we show the visual results of FADM and existing SOTA methods in Fig. \ref{reconstruction fig}. FADM exhibits overall better quality with more fine details than other methods. 


\begin{figure*}[t]
\centering
\includegraphics[width=1\textwidth]{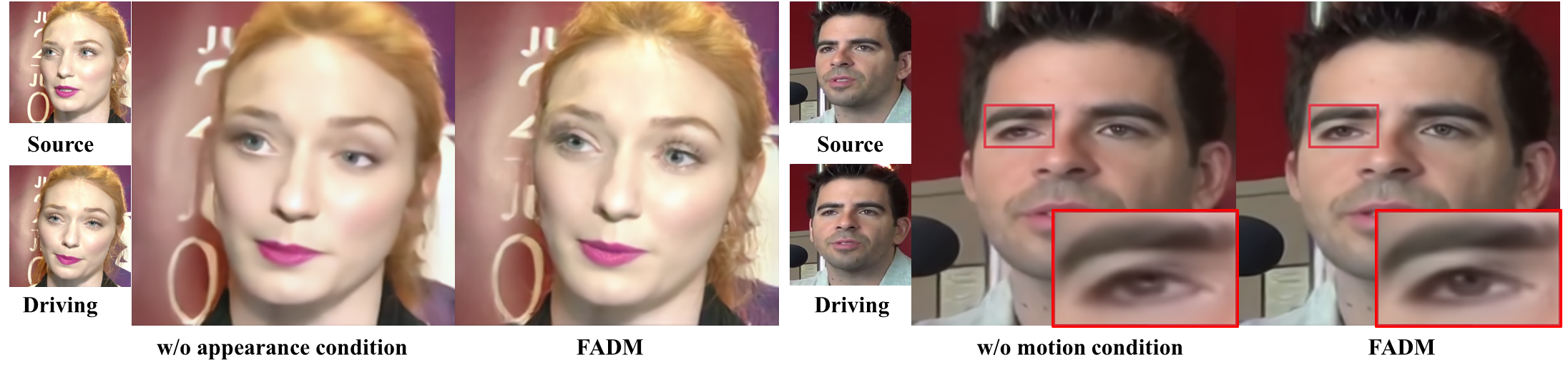} 
\caption{Visualization of the ablation study. Without (w/o) the proposed appearance condition, the animation model performs poorly in synthesizing realistic facial areas (eyes, mouth, and hair). We also specially mark the noteworthy areas in the right part, demonstrating the effectiveness of the designed motion condition in modifying the facial details. }
\label{ablation fig}
\end{figure*}

\begin{table*}[t]
  \caption{Ablation study for same-identity reconstruction on VoxCeleb \cite{nagrani2017voxceleb}.}
  \begin{center}
  \setlength{\tabcolsep}{2mm}{
  \label{ablation}
  \centering
  \renewcommand{\arraystretch}{1.2}
  
  \begin{tabular}{lcccccc}
    \hline
    Method      & $\mathcal{L}_1$ $\downarrow$   & LPIPS $\downarrow$   & PSNR $\uparrow$  & SSIM $\uparrow$  & AKD $\downarrow$  & AED $\downarrow$     \\
    \hline
    w/o color condition       & 0.0428  & 0.1774    & 23.835   & 0.7701   & 1.488  & 0.0263      \\
    w/o motion condition      & 0.0407  & 0.1408   & 24.110   & 0.7818   & 1.400  & 0.0243     \\
    \hline
    FADM           & \textbf{0.0402}   & \textbf{0.1379}  & \textbf{24.434}   & \textbf{0.7841}   & \textbf{1.392}  & \textbf{0.0241}     \\
    \hline
  \end{tabular}
  }
  \end{center}
\vspace{-3mm}
\end{table*}

\paragraph{Cross-Identity Reenactment.}

We validate the effectiveness of FADM on the testing datasets of VoxCeleb, VoxCeleb2, and CelebA for the cross-identity reenactment task, where Face vid2vid \cite{wang2021one} is used to generate the coarse animation results.
Specifically, we randomly select 10 source images and 14 driving videos with different identities from the testing datasets of VoxCeleb and VoxCeleb2 to form various groups. We also use the driving videos of VoxCeleb to animate 10 randomly selected source images from the testing dataset of CelebA. Table \ref{reenactment} shows the quantitative results of the reenactment task. Our FADM outperforms other SOTA methods on the testing datasets of VoxCeleb and CelebA. On VoxCeleb2, FADM exhibits better identity preservation capability, but performs not well in terms of FID. In fact, FID measures the similarity of the data distributions extracted from the two groups. The data quality of VoxCeleb2 is quite poor, including low resolution (the original image size is 224×224) and blurred textures, while FADM tends to generate fine-detailed images, resulting in mismatching between them. Therefore, we think that our FID result on VoxCeleb2 is reasonable.

To prominently show the effectiveness of FADM, we specially select several samples in which the source image and driving videos come from different genders or age groups, and visualize their results in Fig. \ref{reenacment fig}. As we can observe, when the appearance and motion change dramatically, existing SOTA methods may encounter severe distortions or artifacts. In contrast, our FADM can effectively rectify these distortions and enrich the facial details, while ensuring faithful appearance and motion,  thereby generating photo-realistic animation results.

\subsection{Ablation Study}

We conduct comprehensive experiments on the same-identity reconstruction task to demonstrate the effectiveness of the appearance and motion condition mechanism in AGCN, and elaborate why the designed AGCN is the relatively optimal choice for face animation diffusion model over other possible designs. 
\vspace{-1mm}
\paragraph{Appearance Condition.}
In AGCN, we employ an encoder $P_{\operatorname{Conv}}$, which is optimized by $\mathcal{L}_{color}$, to extract the appearance condition from the driving frames during training, and take the coarse animation results as the appearance motion in the inference process. Here we construct another possible design as a comparison model: using the coarse animation results as appearance condition in both training and inference. As shown in the left part of Fig. \ref{ablation fig}, directly using the coarse animation results may fail to synthesize realistic facial regions, 
Moreover, we show the quantitative comparison in Table \ref{ablation}, in which AGCN obtains better results, illustrating the rationality of our appearance condition.

\paragraph{Motion Condition.}
We further evaluate the effectiveness of the motion condition in AGCN.
Specifically, we adopt a comparison model directly using the coarse generated results as the motion condition without the dynamical adjustment according to the 3D reconstruction results in AGCN.
The visualization is shown in the right part of Fig. \ref{ablation fig}.
We can observe that with the designed motion condition, FADM can enrich more fine-grained details, while the comparison model are prone to generate over-smoothing results. Table \ref{ablation} also illustrates that the motion condition in AGCN is effective to improve the quality of generated results.


\section{Conclusion}

In this paper, we propose an attribute-guided Diffusion Model for face animation (FADM), which introduces the iterative diffusion steps to improve the quality of the animation results. To guarantee the generated facial attributes, including appearance and motion, and meet the requirements of face animation, we design an Attribute-Guided Conditioning Network (AGCN) to extract faithful appearance and motion conditions for the subsequent diffusion process. 
Based on the coarse generated results and disentangled poses and expressions predicted by the advanced 3D face reconstruction module, AGCN adaptively modulates the appearance and motion conditions and navigates the diffusion denoising model to synthesize ideal facial details and rectify the distortions.
Moreover, FADM can be directly used to improve the quality of talking-head videos without extra modulation. Extensive experiments demonstrate that our FADM achieves new state-of-the-art performance.


\section{Acknowledgements}
This work was supported by National Natural Science Foundation of China under Grant 62076016, Beijing Natural Science Foundation L223024.

{\small
\bibliographystyle{ieee_fullname}
\bibliography{egbib}
}

\end{document}